\documentclass[journal]{IEEEtran}
\usepackage{amsmath,amsfonts}
\usepackage{algorithmic}
\usepackage{array}
\usepackage[caption=false,font=normalsize,labelfont=rm,textfont=rm]{subfig}
\usepackage{textcomp}
\usepackage{stfloats}
\usepackage{url}
\usepackage{verbatim}
\usepackage{graphicx}
\usepackage{orcidlink}
\usepackage{amssymb}
\usepackage{amsmath}
\usepackage{multirow}
\usepackage{booktabs}
\usepackage{cite}

\hypersetup{
    colorlinks=true,
    linkcolor=black,
    citecolor=black
}

\def\BibTeX{{\rm B\kern-.05em{\sc i\kern-.025em b}\kern-.08em
    T\kern-.1667em\lower.7ex\hbox{E}\kern-.125emX}}
\usepackage{balance}
\usepackage[section]{placeins}
\usepackage{hyperref}
\usepackage{cleveref} 

\begin{document}
\title{Scale-Aware Relay and Scale-Adaptive Loss for Tiny Object Detection in Aerial Images}
\author {
        \IEEEauthorblockN{Jinfu Li\hspace{-1.5mm}$^{~\orcidlink{0009-0008-6013-5675}}$,
         Yuqi Huang\hspace{-1.5mm}$^{~\orcidlink{0009-0007-2482-250X}}$,
         Hong Song\hspace{-1.5mm}$^{~\orcidlink{0000-0002-3171-2604}}$,
         Ting Wang$^{\orcidlink{0009-0000-6638-8130}}$,
         Jianghan Xia$^{\orcidlink{0009-0008-8826-0646}}$,
         Yucong Lin$^{\orcidlink{0000-0002-9039-0318}}$,
         Jingfan Fan$^{\orcidlink{0000-0003-4857-6490}}$,
         Jian Yang\hspace{-0.5mm}$^{\orcidlink{0000-0003-1250-6319}}}$
        \thanks{This work was supported in part by the National Natural Science Foundation of China (No. 62025104, 62171039 and U22A2052) and in part by the Natural Science Foundation of Beijing, China (No. L242024). (Corresponding authors: Hong Song and Jian Yang; E-mail: songhong@bit.edu.cn and jyang@bit.edu.cn.)}
        \thanks{Jinfu Li, Yuqi Huang, Hong Song, Ting Wang and Jianghan Xia are with the School of Computer Science and Technology, Beijing Institute of Technology, Beijing 100081, China.}
        \thanks{Yucong Lin, Jingfan Fan and Jian Yang are with the School of Optics and Photonics, Beijing Institute of Technology, Beijing 100081, China.}
}

\maketitle

\begin{abstract}
Recently, despite the remarkable advancements in object detection, modern detectors still struggle to detect tiny objects in aerial images. One key reason is that tiny objects carry limited features that are inevitably degraded or lost during long-distance network propagation. Another is that smaller objects receive disproportionately greater regression penalties than larger ones during training. To tackle these issues, we propose a Scale-Aware Relay Layer (SARL) and a Scale-Adaptive Loss (SAL) for tiny object detection, both of which are seamlessly compatible with the top-performing frameworks. Specifically, SARL employs a cross-scale spatial-channel attention to progressively enrich the meaningful features of each layer and strengthen the cross-layer feature sharing. SAL reshapes the vanilla IoU-based losses so as to dynamically assign lower weights to larger objects. This loss is able to focus training on tiny objects while reducing the influence on large objects. Extensive experiments are conducted on three benchmarks (\textit{i.e.,} AI-TOD, DOTA-v2.0 and VisDrone2019), and the results demonstrate that the proposed method boosts the generalization ability by 5.5\% Average Precision (AP) when embedded in YOLOv5 (anchor-based) and YOLOx (anchor-free) baselines. Moreover, it also promotes the robust performance with 29.0\% AP on the real-world noisy dataset (\textit{i.e.,} AI-TOD-v2.0).

\end{abstract}


\section{Introduction}\label{intro}

\IEEEPARstart{T}{iny} object detection (TOD) is always a fundamental task in aerial image processing, aiming to categorize and locate highly structured objects with fewer than 16$\times$16 pixels, such as persons, ships and vehicles. These tiny objects appear everywhere in practical applications, where their accurate detection is beneficial to automated interpretation and informed decision-making \cite{OSCO20211,2024_MARFPNet,2023_YOLOv5s-M,ZHENG2021112636}. However, the shooting angle and imaging altitude enable the size, shape, and orientation of objects to vary so much that extracting sufficient appearance information is extremely difficult \cite{2022Tongsurvey}. Meanwhile, changes in light and weather further complicate the backgrounds, leading to a low signal-to-noise ratio (SNR) \cite{2023_metareview}. The cumulative effects of these factors pose significant challenges for TOD, making it a hotspot in both academia and industry circles.
\begin{figure}[h]
  \centering
  \includegraphics[width=\linewidth]{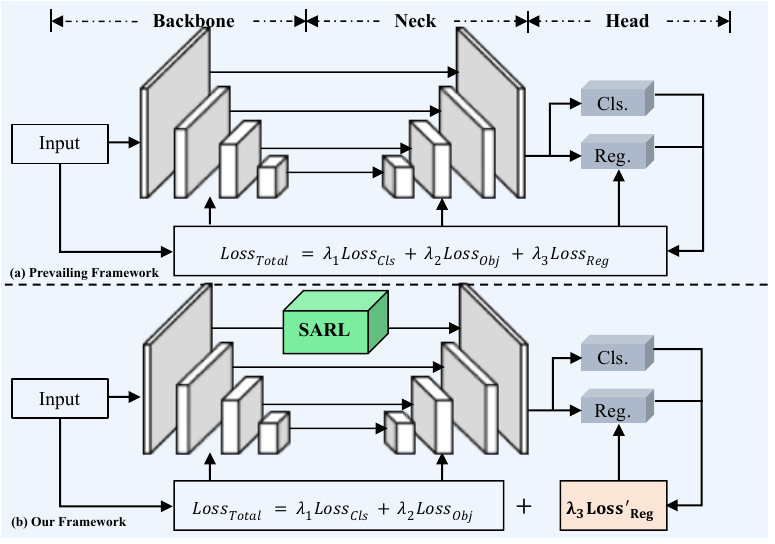}
  \caption{Schematic illustration of the comparison between the prevailing framework and our framework. (a) Prevailing Framework. (b) Our Framework.}
  \label{fig:fig1}
\end{figure}

With the rapid development of deep neural networks \cite{FasterRCNN2017,CascadeRCNN2018,2020DETR}, generic object detection has made great progress. Many studies suggested that general detectors, even the most advanced ones, will not produce satisfactory results if directly applied to the TOD task \cite{AI_TOD2020}. The alternatives usually assemble some special design and/or mechanisms into the top of general detectors. A common strategy is to increase the number of tiny objects in the dataset, thereby enhancing the model's ability to learn their representations. These data augmentation techniques include Generative adversarial networks (GAN)-based object synthesis\cite{bosquet2023full}, image pyramid \cite{shamsolmoali2021}, aggregated-mosaic \cite{rs13132602}, copy-paste \cite{XIAO2023118665}, and so on. However, these methods entail a compromise in computational costs. Given the scarcity of visual cues, it is necessary to mine and exploit extra semantic and contextual information. For instance, Cheng \textit{et al.} \cite{cheng2022tiny} employed cross-attention and self-attention network (RCSANet) to match key points with regional backgrounds. Chen \textit{et al.} \cite{chen2023mdct} utilized multiple kernel dilated convolution to expand the receptive field without reducing the spatial resolution. Moreover, a lot of efforts have been made on bottom-up and top-down structures \cite{hong2021sspnet,2020IJSTAR,2022Context-Aware}. Recently, researchers began exploring new similarity metrics to replace IoU in NMS (Non-Maximum Suppression), loss functions, and label assignment \cite{2019_sigNMS,Xu2022_NWD,RFLA2022,zhang2023inneriou}. In \cite{2019_sigNMS}, the Sig-NMS is proposed to adjust the threshold by a fraction reset function, which mitigates the problem of tiny objects being suppressed in original NMS. Xu \textit{et al.} \cite{Xu2022_NWD} projected high-dimensional features into 2D Gaussian distributions and adopted the normalized Wasserstein distance (NWD) to represent the location relationship. Later, Xu \textit{et al.} \cite{RFLA2022} also proposed a receptive field-based label assignment (RFLA) strategy, which took a Gaussian receptive field distance to calculate the similarity between predictions and ground truths. While the preceding methods have shown promise, there is still plenty of room for improvement on two issues.

({\romannumeral1}) \textit{Discriminative features degrade or get lost as they propagate through the detection network.} Current mainstream detectors typically consist of three key parts: the backbone, neck, and head. The backbone mainly employs a cascade of convolutional and pooling layers to generate hierarchical feature maps. As the layers deepen, these maps increasingly contain semantic information, which helps recognize objects with diverse scales and contexts. However, this downsampling procedure inevitably reduces the spatial resolution and discards spatial details that are crucial for distinguishing tiny objects from their backgrounds and from one another. Additionally, the neck (\textit{e.g.}, FPN \cite{lin2017fpn}, PANet \cite{liu2018panet}, and BiFPN \cite{2020bifpn}) directly utilizes a series of operations like upsampling, concatenation, and skip/lateral connections to aggregate low-layer and high-layer feature maps. Ideally, it should work in harmony with the backbone, with each part leveraging its unique strengths to build a comprehensive feature pyramid. However, the neck largely relies on the quality and completeness of the input to be effective. Weak or incomplete feature maps prevent the neck from refining them, causing a cascading effect that exacerbates the performance of the entire detection network.

({\romannumeral2}) \textit{Unfair regression penalties are imposed on tiny objects in bounding box predictions.} As we all know, IoU is the most familiar metric for evaluating the accuracy of predictions. The regression losses based on IoU and its derivatives (\textit{e.g.}, GIoU\cite{2019GIOU}, DIoU and CIoU\cite{2020DIOU}) have been proven effective in guiding the model to learn precise bounding box coordinates. Nevertheless, when these losses are implemented in TOD, a significant disparity in penalty arises. Concretely, large objects, due to larger pixels areas, exhibit a higher tolerance for deviations in localization without sharp drops in IoU values. This means that the model prioritizes optimizing its performance on large objects since they contribute less to the regression loss. In contrast, tiny objects have a greater sensitivity to localization deviations. Even minor positional shifts can decay IoU values rapidly. As a result, these values translate into disproportionate penalties in the calculation of the regression loss, focusing the model less on tiny objects.

To mitigate the above issues, this paper introduces a Scale-Aware Relay Layer (SARL) and a Scale-Aware Loss (SAL) for tiny object detection. Both can be smoothly integrated into previous methods. SARL incorporates a cross-scale spatial-channel attention mechanism that refines and enhances the hierarchical feature maps extracted by the backbone before feeding them into the neck. This mechanism exploits high-level semantic information to reweight low-level spatial information from adjacent layers. By dong so, SARL preserves and amplifies the fine-grained details that are often weakened or lost during network traversal while suppressing redundant information. Additionally, SAL is a dynamically scaled IoU-based regression loss, where the scaling factor decreases fast as the object's area increases. In fact, this scaling factor automatically reduces the penalty of larger objects during training, enabling the model focus more on tiny objects. To validate SARL and SAL, we conducted extensive experiments on four public datasets and compared their performance when integrated with anchor-based (YOLOv5) and anchor-free (YOLOx) object detection frameworks. Our results demonstrate the effectiveness of this approach in improving the overall performance, particularly in challenging noise scenarios.

The main contributions are summarized as follows:
\begin{itemize}
  \item{We design the scale-aware relay layer (SARL), positioned between the backbone and neck, to selectively emphasize the regions of interest while suppressing irrelevant features, thereby preventing the loss of fine-grained details.}
  \item{We design the scale-adaptive loss (SAL), which dynamically assigns regression penalties to objects based on their scale, ensuring a robust training against scale variations.}
  \item{We incorporate SARL and SAL into the popular anchor-based YOLOv5 and anchor-free YOLOx object detection algorithms, respectively, and demonstrate their performance improvement on four major benchmarks.}
\end{itemize}


The rest of this paper is organized as follows. In \autoref{Related Work}, we provide a brief overview of related work in object detection and tiny object detection. In \autoref{Method}, we describe the architecture and design of the SARL and SAL in detail. In \autoref{Experiment}, we present our experimental setup and results, including a comparison with state-of-the-art detection models. Finally, in \autoref{Conclusion}, we conclude our findings and discuss potential future directions for research in this area.

\section{Related Work}\label{Related Work}
\subsection{Generic Object Detection}
Object detection methods can be roughly grouped into anchor-based and anchor-free detectors, depending on whether the pre-defined anchor boxes are used. The former utilizes one-stage or two-stage networks to generate region proposals and perform classification. Instead, the latter directly predicts the bounding boxes by keypoint-based or center-based networks.
\subsubsection{Anchor-based Detectors}
One-stage networks, such as SSD\cite{SSD2016}, RetinaNet\cite{RetinaNet2020}, and YOLO family\cite{Yolov32018,YOLOv42020,2022yolov5}, prioritized speed and real-time performance. They accomplished this by classifying and regressing bounding boxes in a single pass through the network, making them ideal for applications where latency is a critical factor. In contrast, two-stage networks, exemplified by FPN\cite{lin2017fpn}, R-CNN family\cite{RCNN2014,FastRCNN2015,FasterRCNN2017,CascadeRCNN2018}, and TridentNet\cite{TridentNet2019}, emphasized higher accuracy by first creating a set of region proposals and then refining these proposals through a second stage of classification and bounding box regression. This two-step process allowed for more precise localization and classification of objects, making two-stage networks effective in scenarios requiring high accuracy.
\subsubsection{Anchor-free Detectors}
Keypoint-based networks inferred bounding boxes from the geometrical relationships of multiple key points. CornerNet\cite{CornerNet2018} detected the top-left and bottom-right corners with embedding vectors, while CenterNet\cite{zhou2019objects} added the center point into the detection. Other representative works included Grid R-CNN\cite{2019grid}, RepPoints\cite{Reppoints2019}, and FoveaBox\cite{2020foveabox}. In contrast, center-based networks mainly identified the center point. FCOS\cite{FCOS2020} computed distances from the center to the box boundaries, and TOOD\cite{feng2021tood} optimized anchor points via a task alignment learning strategy. Meanwhile, FCOS and TOOD have also introduced into the YOLO series, \textit{e.g.}, YOLOx\cite{YOLOX2021} and YOLOv8 \cite{2023yolov8}. Obviously, anchor-free detectors offered a compelling alternative, especially in resource-limited or highly variable detection tasks.
\subsection{Tiny Object Detection}
\subsubsection{Discriminative Feature Learning}
It is well-known that tiny objects have limited appearance information. Researchers have therefore investigated various methods to improve the learning of discriminative features for them. In \cite{2020IJSTAR}, Li \textit{et al.} proposed a cross-layer attention mechanism after downsampling and upsampling procedures to strengthen the expression of spatial and context information. In \cite{2022msodanet,2022Context-Aware}, the authors tried to establish the connection between different level features and combine them by a bidirectional FPN. Wu \textit{et al.}\cite{2022FSANet} introduced a feature-and-spatial aligned network (FSANet) where a novel feature-aware alignment module is designed to align adjacent features of different resolutions, facilitating the extraction of more discriminative features. Like RCSANet\cite{cheng2022tiny}, FSANet also followed an anchor-free paradigm that could reduce the position-sensitive influences on tiny objects. Besides, several early mentioned works \cite{hong2021sspnet,chen2023mdct} on hierarchical feature learning and interaction have shown impressive performance. However, these methods preferred to design suitable network structures to enrich discriminative features, overlooking the coordination with the training strategies.
\subsubsection{Improved Detection Metrics}
As stated in \autoref{intro}, several excellent studies (\textit{e.g.}, NWD\cite{Xu2022_NWD} and RFLA\cite{RFLA2022}) have highlighted the challenges associated with most IoU-based detection metrics in handling position deviation and delineating object boundaries. Thus, there is a trend towards creating specialized metrics for NMS, loss functions, and label assignment processes\cite{DotD2021,KLDetTGRS2024,rs16234485}. In their research, Xu \textit{et al.}\cite{DotD2021} introduced the Dot Distance (DotD), which is defined as the normalized Euclidean distance between the center points of two bounding boxes. Zhou \textit{et al.}\cite{KLDetTGRS2024} proposed the use of the Kullback-Leibler divergence (KLD) as a replacement for the NWD\cite{Xu2022_NWD} to select more positive instances of tiny objects. In \cite{rs16234485}, Su \textit{et al.} firstly modelled bounding boxes as 2D Gaussian distributions similar to \cite{Xu2022_NWD} and \cite{KLDetTGRS2024}. They then designed a new metric called Mixed Minimum Point-Wasserstein (MMPW) to analyze these Gaussian distributions rather than NWD and KLD. However, the preceding methods mainly concentrated on optimizing training strategies, while lacking in-depth exploration of discriminative features.

In summary, both approaches have their respective focuses yet fall short in achieving optimal accuracy and generalization. Therefore, this paper proposes the Scale-Aware Relay Layer (SARL) and Scale-Adaptive Loss (SAL) to merge their merits and enhance model performance. SARL dynamically adjusts information flow within the network based on input feature scale, ensuring effective propagation of relevant features through a cross-layer attention mechanism that considers both semantic and spatial information. Additionally, SAL complements SARL by modulating error penalties based on object scale, enabling the model to pay more attention on smaller objects while maintaining robustness for larger ones.



\section{Proposed Method}\label{Method}
In this section, we describe our method in detail. First, Section \ref{sec1} introduces the overview of the proposed framework. Then, Section \ref{sec2} presents the Scale-Aware Attention Network ($SA^2N$). Finally, the Scale-Feedback Loss ($SFL$)  is introduced in Section \ref{sec3}.

\subsection{Overall Framework} \label{sec1}

\begin{figure*}[!htbp]
  \centering
  \includegraphics[width=\linewidth]{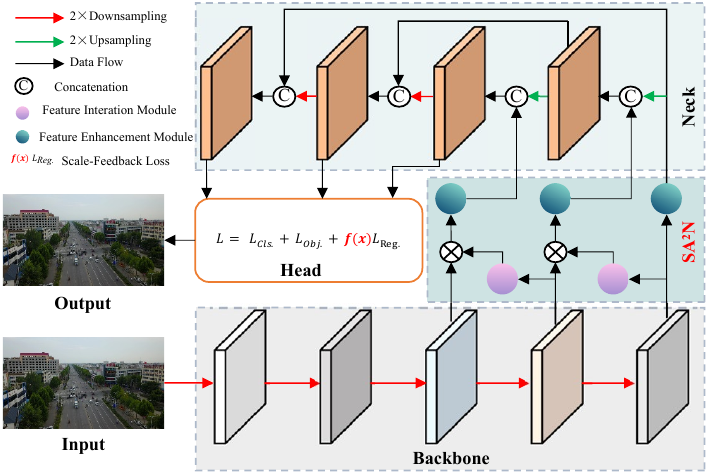}
  \caption{Overall framework of our method. $SA^2N$ is used to enhance discriminative features. $SFL$ is used to  balance the regression loss.}
  \label{fig:fig2}
\end{figure*}
Before discussing the particular architecture as shown in \ref{fig:fig2}, we should clarify the main design ideal first. The current object detection paradigms faces two main challenges in effectively identifying small objects in aerial images, namely the degradation of discriminative features during feature aggregation and the significant imbalance in regression loss of objects of different scales. To this end, we propose a novel approach that combines mix-attention mechanism with scale-aware optimization strategy to improve the quality of discriminative features and balance the regression loss across various scales.

\ref{fig:fig1} shows a comparison between our framework and prevailing frameworks. From \ref{fig:fig2}, it can be seen that $SA^2N$ is one of our main innovations. This work aims to address a critical bottleneck: the insufficiency or loss of discriminative features during pyramid feature aggregation processes. Existing feature extraction paradigms often fail to provide adequate multi-level discriminative features, which is crucial for accurate tiny object detection. This problem is exacerbated by the downsampling and pooling processes in the network's backbone, resulting in blurred appearances of the targets and their low signal-to-noise ratios. To mitigate these issues, our approach introduces the Self-Attention Aggregated Network ($SA^2N$). $SA^2N$ leverages attention mechanisms to adaptively weight the pyramidal feature hierarchy extracted from the backbone. By doing so, it enhances the quality of multi-scale features, facilitating effective bidirectional aggregation and significantly improving the network's capability to detect small objects in complex aerial imagery environments.

Moreover, as shown in \ref{fig:fig2}, our method incorporates the Scale-Feedback Loss ($SFL$), which uses scale-feedback signals to calculate localization errors, thereby promoting balanced network training. The $SFL$ mechanism ensures that the regression loss is distributed evenly across various scales, which is crucial for achieving consistent detection performance. This balance addresses another critical issue in tiny object detection, where different scales can lead to inconsistent loss contributions.

Both $SA^2N$ and $SFL$ are designed to be compatible with a wide range of detection frameworks, including both anchor-based and anchor-free detectors that utilize IoU-based regression loss. This compatibility provides flexibility and robustness, making our approach versatile for different application scenarios. $SA^2N$'s ability to extract precise dense features is pivotal in improving detection accuracy, particularly for small objects. The combination of a mix-attention mechanism with a scale-aware optimization strategy in our novel approach significantly advances object detection technologies. It ensures that even in the presence of complex aerial imagery, tiny objects can be detected with high precision and reliability.

\subsection{Scale-Aware Attention Block} \label{sec2}

\begin{figure}[!htbp]
  \centering
  \includegraphics[width=\linewidth]{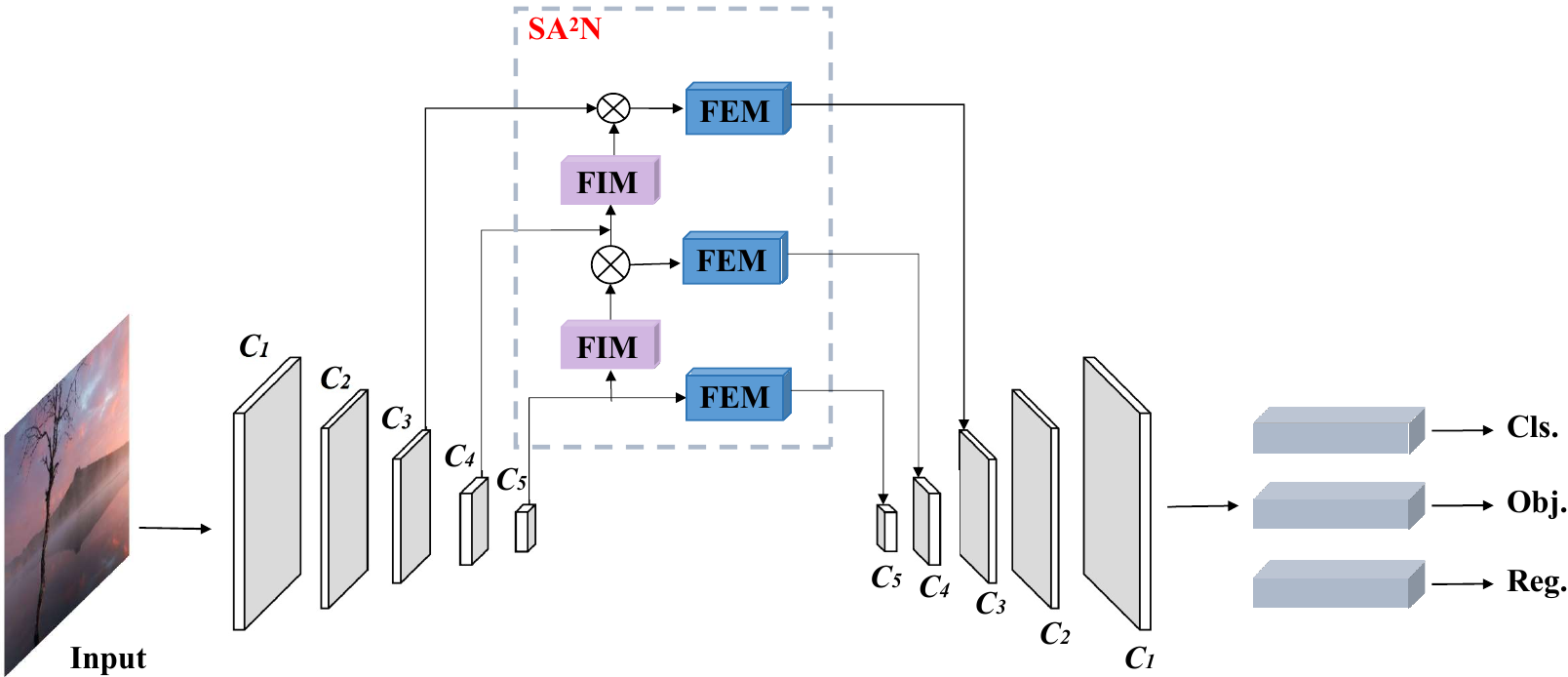}
  \caption{Position and overall structure of $SA^2N$.}
  \label{fig:fig3}
\end{figure}

\begin{figure}[!htbp]
  \centering
  \includegraphics[width=\linewidth]{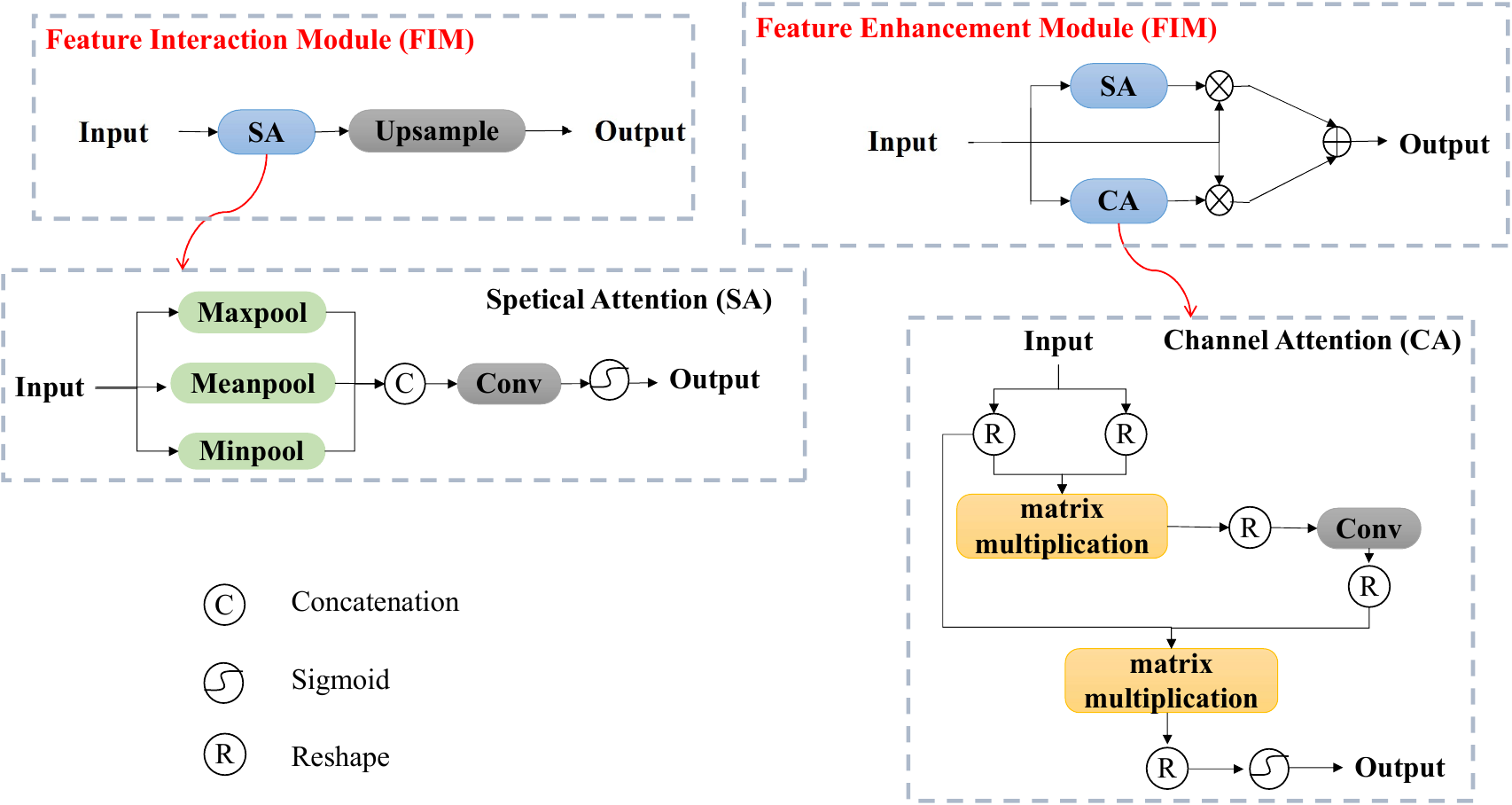}
  \caption{Specific structure of $FIM$ and $FEM$.}
  \label{fig:fig4}
\end{figure}
\subsection{Scale-Feedback Loss} \label{sec3}
In common with other object detection methods, the loss function of our framework takes the form of multi-task loss, which is defined as
\begin{equation}
\label{total_loss}
Loss = Loss_{cls}+Loss_{obj}+Loss_{pos}
\end{equation}
where $Loss_{cls}$, $Loss_{obj}$, and $Loss_{pos}$ represent classification loss, confidence loss, and positional loss, respectively.
Both $Loss_{cls}$ and $Loss_{obj}$ use the cross-entropy loss (CE). It can be formulated in the following form.
\begin{equation}
\label{CE_Loss}
Loss_{cen} = -\frac{1}{N}\sum_{i=1}^{n}[a_{i}\log\sigma(a^{'}_{i})+(1-a_{i})log(1-\sigma(a^{'}_{i}))]
\end{equation}
where $Loss_{cen}$ denotes the CE loss. $a^{'}_{i} $represents the predicted value of the model: for classification loss, it is the predicted classification score of prediction box $i$, and for confidence loss, it is the predicted confidence score of the model on the likelihood of the object in the corresponding grid of prediction box $i$. $a_{i}$ represents the true label: for classification loss, it is the true category label of the object, and for confidence loss, it is the true score of whether the object exists in the corresponding grid of prediction box $i$.

In order to alleviate the imbalance of regression loss, we use $SFL$ to replace the original IoU Loss as part of the position loss. The position loss of our model  is as follows (The position loss is a combination of SFL and L1 loss, it can be formulated in the following form.):
\begin{equation}
\label{pos_Loss}
Loss_{pos} = Loss_{l1} + \alpha Loss_{SFL}
\end{equation}
where $\alpha$ is a regulatory factor used to regulate the ratio of L1 loss and $SFL$. L1 loss can be formulated in the following form.
\begin{equation}
\label{L1_Loss}
\begin{split}
&Loss_{l1} = \frac{\sum_{i=1}^{n}|a^{'}_{i}-a_{i}|}{n}\\
&(a_{i}=x_{i},y_{i},h_{i},w_{i}; a^{'}_{i}=x^{'}_{i},y^{'}_{i},h^{'}_{i},w^{'}_{i})
\end{split}
\end{equation}
where $x_{i}$, $y_{i}$, $h_{i}$, $w_{i}$ respectively represent the top left horizontal and vertical coordinates, width and height of the ground truth i, and $x^{'}_{i}$, $y^{'}_{i}$, $h^{'}_{i}$, $w^{'}_{i}$ respectively represent the top left horizontal and vertical coordinates, width and height of the predicted box $i$. $a_{i}$ and $a^{'}_{i} $ are respectively substituted with $x_{i}$, $y_{i}$, $h_{i}$, $w_{i}$ and $x^{'}_{i}$, $y^{'}_{i}$, $h^{'}_{i}$, $w^{'}_{i}$, and the L1 loss of the horizontal and vertical coordinates, width, and height of the top left corner of the prediction box $i$ can be calculated.

$SFL$ can be formulated in the following form.
\begin{equation}
\label{SFL}
\begin{split}
Loss_{SFL}&=\sum_{i=1}^{n}\beta \ln({2-s_{i}})(1-IoU_{i}^{2})\\
s_{i}&=\frac{h_{i}w_{i}-\min_{i=1}^{n}h_{i}w_{i}}{\max_{i=1}^{n}h_{i}w_{i}-\min_{i=1}^{n}h_{i}w_{i}}\\
IoU_{i}&=\frac{(x^{'}_{i}+w^{'}_{i}-x_{i})(y^{'}_{i}+h^{'}_{i}-y_{i})}{h^{'}_{i}w^{'}_{i}+h_{i}w_{i}-(x^{'}_{i}+w^{'}_{i}-x_{i})(y^{'}_{i}+h^{'}_{i}-y_{i})}
\end{split}
\end{equation}
where $s_{i}$ represents the area of ground truth i after normalization, $IoU_{i}$ represents the ratio of the intersection area of predicted box i and ground truth i to the area of the disease. $\ln{(2-s_{i})}$ is an adjustment factor that enables the network to adaptively adjust the position loss of the object based on its size.

\textbf{!!!DATA!!}\textbf{As shown in the figure, the imbalance of regression loss is mainly reflected in larger objects receiving higher penalties compared to smaller objects.} Therefore, $SFL$ mainly suppresses this imbalance. It is not difficult to find that $\ln{(2-s_{i})}$ is a subtraction function with the predicted box size as the independent variable. Under the influence of this regulatory factor, the smaller the object size, the greater the increase in the proportion of object position loss. Conversely, the larger the object size, the greater the decrease in object position loss. In this way, the network will have more sufficient monitoring signals for small objects, and the training of the object detection network will be more balanced. $SFL$ is easily integrated into any detection framework, with almost no additional burden during training and inference. The additional cost only comes from the loss statistics calculation, which can be almost negligible compared to the more computationally intensive forward and backward propagation. $\beta$ is a hyperparameter used to control the magnitude of the regulatory factor's effect on position loss.

\section{Experiments}\label{Experiment}
\subsection{Datasets}
\noindent {\bf{AI-TOD:}} The AI-TOD dataset \cite{AI_TOD2020} collects $28,036$ images from five publicly avaiable and large-scale remote sensing datasets. It is specifically tailored for the challenging task of tiny object detection in aerial imagery. In AI-TOD, $700,621$ object instances are meticulously annotated with an average size of just 12.8 pixels. These instances are categorized into eight distinct classes: airplane (AI), bridge (BR), storage-tank (ST), ship (SH), swimming-pool (SP), vehicle (VE), person (PE), and wind-mill (WM). Moreover, the objects are further classified based on their absolute sizes into very tiny ($2\thicksim8$ pixels), tiny ($8\thicksim16$ pixels), small ($16\thicksim32$ pixels), and medium ($32\thicksim64$ pixels) groups, with no large objects included. There are $40\% (11,214)$, $10\% (2,804)$, and $50\% (14,018)$ images for training, validation, and testing.

\noindent {\bf{AI-TOD-v2:}} It is an upgraded version of original AI-TOD dataset \cite{AI_TOD2020}.

\noindent {\bf{VisDrone2019:}}  \cite{zhu2021detection}

Hyphens and periods should not be used in equation numbers, i.e., use (1a) rather than
(1-a) and (2a) rather than (2.a) for sub-equations. This should be consistent throughout the article.

\noindent {\bf{DOTA-v2.0:}}

\subsection{Implementation Details}
All the experiments in this paper are conducted on two RTX 3080Ti GPUs with the batch size of 4.

\subsection{Evaluation Metrics}
The Microsoft COCO benchmark \cite{2014coco} is widely used for evaluating object detection models, owing to its fine-grained assessment of a model's ability across varying IoU thresholds. Following this standard, we employ the average precision (AP), $AP_{0.5}$, and $AP_{0.75}$ as quantitative measures to evaluate the detection performance. Furthermore, TOD focuses on tiny objects with smaller than $16\times16$ pixels \cite{AI_TOD2020}, while COCO defines small objects that occupy an area less than $32\times32$ pixels. To accurately reflect our model's performance, we adopt $AP_{vt}$ (very tiny), $AP_{t}$ (tiny), $AP_{s}$ (small), and $AP_{m}$ (medium) as additional metrics.
  
\subsection{Comparative Experiments}
  
  \begin{table*}[hbtp]
  \center
  \caption{Quantitative Results (i.e., Percentage of AP Scores) of Our Detector with Some Representative Detectors on AI-TOD Dataset. The Best Result is Highlighted in \textbf{Bold}. `NG' Indicates that the Original Publication Did not Give the Corresponding Values.}
  \label{tab:tabel1}         
  \renewcommand{\arraystretch}{1.5}
  \resizebox{\textwidth}{!}{
  \begin{tabular}{cccccccccccccccccc}
  \toprule
  Methods  &Venue &\multicolumn{1}{c|}{Backbone} & $AP$ & $AP_{0.5}$ &\multicolumn{1}{c|}{$AP_{0.75}$} & $AP_{vt}$ & $AP_t$ & $AP_s$ & \multicolumn{1}{c|}{$AP_m$} &AI	&BR	&ST	&SH	&SP	&VE	&PE	&WM \\ \midrule
  \textbf{Anchor-based Detectors:} \\
  RetinaNet~\cite{RetinaNet2020} &ICCV'2017 & \multicolumn{1}{c|}{ResNet-50}  & 8.7 & 22.3  & 4.8  & 2.4  & 8.9  & 12.2  & \multicolumn{1}{c|}{16.0}  &0.0 	&6.6 	&1.8 	&20.9 	&0.1 	&5.7 	&1.8 	&0.5\\
  TridentNet~\cite{TridentNet2019} &ICCV'2019 & \multicolumn{1}{c|}{ResNet-50} & 7.5  & 20.9  & 3.6  & 1.0  & 5.8  & 12.6  &\multicolumn{1}{c|}{14.0}   &9.7 &0.8 	&12.3 	&17.1 	&3.2 	&11.9 	&4.0 	&0.9 \\
  Faster R-CNN~\cite{FasterRCNN2017} &TPAMI'2017 & \multicolumn{1}{c|}{ResNet-50} &11.4 	&27.0 	&8.0 	&0.0 	&8.3 	&23.1 	&\multicolumn{1}{c|}{24.5} 	&22.7 	&3.9 	&20.2 	&19.0 	&8.9 	&11.9 	&4.5 	&0.3 \\
  YOLOv5~\cite{2022yolov5}		&ArXiv'2022 & \multicolumn{1}{c|}{CSP-DarkNet}	&11.9 	&28.5 	&7.6 	&3.1 	&12.0 	&14.2 	&\multicolumn{1}{c|}{20.3} 	&9.3 	&6.4 	&20.3 	&32.9 	&0.9 	&19.3 	&5.3 	&0.7 \\
  Cascade R-CNN~\cite{CascadeRCNN2018} &CVPR'2018 & \multicolumn{1}{c|}{ResNet-50} & 13.8  & 30.8  & 10.5  & 0.0  & 10.6  & 25.5  &\multicolumn{1}{c|}{26.6}   &25.6 	&7.5 	&23.3 	&23.6 	&10.8 	&14.1 	&5.3 	&0.0 \\
  DoTD \cite{DotD2021}	&CVPRW'2021	& \multicolumn{1}{c|}{ResNet-50}	&16.1 	&39.2 	&10.6 	&8.3 	&17.6 	&18.1 	&\multicolumn{1}{c|}{22.1} &\multicolumn{8}{c}{NG}\\							
  CFINet \cite{CFINet2023}	&ICCV'2023  &\multicolumn{1}{c|}{ResNet-50}	&20.1 	&49.1 	&11.3 	&3.8 &20.7 	&25.9 	&\multicolumn{1}{c|}{36.6} 	&40.0 	&6.8 	&35.5 	&33.5 	&15.7 	&20.9 	&7.9 	&0.6 \\
  DINO-Deformable-DETR \cite{zhang2023dino}	&ICLR'2023	&\multicolumn{1}{c|}{ResNet-50}	&23.2 	&56.6 	&15.4 	&9.9 	&23.1 	&29.3 	&\multicolumn{1}{c|}{37.6} &\multicolumn{8}{c}{NG}\\	
  NWD-RKA\cite{Xu2022_NWD} &ISPRS'2022  &\multicolumn{1}{c|}{ResNet-50} &23.4 	&53.5 	&16.8 	&8.7 	&23.8 	&28.5 	&\multicolumn{1}{c|}{36.0} &\multicolumn{8}{c}{NG}\\
  RFLA~\cite{RFLA2022}   &ECCV'2022  &\multicolumn{1}{c|}{ResNet-50}  &24.8 &55.2  &18.5  &9.3  &24.8  &30.3  &\multicolumn{1}{c|}{38.2}  &\multicolumn{8}{c}{NG} \\
  DNTR \cite{2024DNTR}	&TGRS'2024	&\multicolumn{1}{c|}{ResNet-50}	&26.2 	&56.7 	&20.2 	&12.8 	&26.4 	&31.0 	&\multicolumn{1}{c|}{37.0} &\multicolumn{8}{c}{NG}\\ \midrule
  \textbf{Anchor-free Detectors:} \\
  FoveaBox~\cite{2020foveabox} &TIP'2020  &\multicolumn{1}{c|}{ResNet-50}	&8.1 	&19.8 	&5.1 	&0.9 	&5.8 	&13.4 	&\multicolumn{1}{c|}{15.9} 	&13.8 	&0.0 	&18.5 	&17.7 	&0.0 	&11.4 	&3.4 &0.0\\
  RepPoints~\cite{Reppoints2019}  &ICCV'2019  &\multicolumn{1}{c|}{ResNet-50}  & 9.2 & 23.6  & 5.3  & 2.5  & 9.2  & 12.9  &\multicolumn{1}{c|}{14.4}  &2.9 	&2.3 	&21.4 	&26.4 	&0.0 	&15.2 	&5.4 	&0.0  \\
  CornerNet~\cite{CornerNet2018}	&ECCV'2018	&\multicolumn{1}{c|}{Hourglass-104}	&9.0 	&25.4 	&4.3 	&2.9 	&12.9 	&11.5 	&\multicolumn{1}{c|}{7.0} 	&10.6 	&11.8 	&14.1 	&16.9 	&4.4 	&7.0 	&4.8 	&2.7\\ 
  Grid R-CNN~\cite{2019grid}	&CVPR'2019	&\multicolumn{1}{c|}{ResNet-50}	&12.2 	&27.7 	&9.0 	&0.2 	&10.3 	&22.6 	&\multicolumn{1}{c|}{23.3} 	&22.6 	&8.6 	&18.9 	&22.0 	&7.3 	&12.9 	&4.8 	&0.4\\ 
  CenterNet~\cite{zhou2019objects}	&ICCV'2019	&\multicolumn{1}{c|}{DLA-34}	&13.4 	&39.2 	&5.0 	&3.8 	&12.1 	&17.7 	&\multicolumn{1}{c|}{18.9} 	&17.4 	&9.5 	&25.9 	&21.9 	&6.2 	&16.5 	&8.1 	&1.9\\
  FCOS~\cite{FCOS2020} &ICCV'2019  &\multicolumn{1}{c|}{ResNet-50}  & 12.6 & 30.4  & 8.1  & 2.3  & 12.2  & 17.2  &\multicolumn{1}{c|}{25.0} &19.3 	&7.4 	&26.1 	&27.4 	&5.8 	&17.4 	&5.6 	&2.6\\ 
  M-CenterNet~\cite{AI_TOD2020}  &ICPR'2021  &\multicolumn{1}{c|}{DLA-34}  & 14.5 & 40.7  & 6.4  & 6.1  & 15.0  & 19.4  &\multicolumn{1}{c|}{20.4}   &18.6 	&10.6 	&27.6 &22.3 	&7.5 	&18.6 	&9.2 	&2.0 \\ 
  YOLOv8 \cite{2023yolov8}	&ArXiv'2023	&\multicolumn{1}{c|}{ResNet-50}	&14.9 	&32.5 	&11.6 	&4.5 	&14.4 	&19.0 	&\multicolumn{1}{c|}{29.3} &\multicolumn{8}{c}{NG}\\
  YOLOx~\cite{YOLOX2021}  &ArXiv'2021 &\multicolumn{1}{c|}{CSPDarknet-53} & 21.1 & 48.6  & 14.7 & 7.7  & 21.4 & 28.9 &\multicolumn{1}{c|}{33.9} &33.8 	&12.2 	&37.1 	&33.6 	&15.9 	&27.3 	&13.5 	&4.2  \\
  FSANet~\cite{2022FSANet}	&TGRS'2022 &\multicolumn{1}{c|}{Swin-T}	&22.6 	&52.8 	&15.6 	&7.4 	&21.6 	&29.1 	&\multicolumn{1}{c|}{38.5} 	&30.9 	&15.1 	&35.0 	&40.3 	&19.8 	&24.9 &8.9 	&5.6\\  \midrule
  \textbf{Ours:} \\
  YOLOx w/ $SA^2B$-SAL &- &\multicolumn{1}{c|}{CSPDarknet-53} & 26.6 & 59.1 & 20.8 & 10.0 & 26.0 & 33.7 &\multicolumn{1}{c|}{39.9} &38.9	&16.5	&40.3	&40.8	&22.1	&29.7	&16.9	&7.9\\
  YOLOv5 w/ $SA^2B$-SAL &- &\multicolumn{1}{c|}{CSPDarknet-53} &\textbf{29.6} 	&\textbf{62.6} 	&\textbf{24.1} 	&\textbf{11.4} 	&\textbf{28.7} 	&\textbf{38.6} 	&\multicolumn{1}{c|}{\textbf{47.4}} 	&\textbf{44.5} 	&\textbf{19.7} 	&\textbf{43.3} 	&\textbf{42.0} 	&\textbf{26.9} 	&\textbf{33.7} 	&\textbf{16.2} 	&\textbf{10.6} 
  \\ \bottomrule
  \end{tabular}}
  \end{table*}

  \subsection{Generalization Experiments}
  It is necessary to evaluate the generalization capability of the proposed method in different aerial scenarios. Therefore, we select one 
  \begin{table}
  \center
  \caption{Comparison Results on VisDrone2019 and DOTA-v2.0 Validation Sets. `*' Denotes the Detector with $SA^2B$ and SFL.}
  \label{tab:tabel7}    
  \renewcommand{\arraystretch}{1.5}
  \resizebox{\linewidth}{!}{
    \begin{tabular}{cc|ccc|cccc}\toprule
      Datasets & Methods & $AP$  & $AP_{50}$  & $AP_{75}$  & $AP_{vt}$  & $AP_t$  & $AP_s$  & $AP_m$ \\ \midrule
      \multirow{2}{*}{VisDrone2019} & YOLOx             & 23.9 & 40.2 & 24.1 & 4.3 & 10.6 & 21.2 & 34.7\\ 
                                    & YOLOx*  & $\mathbf{26.1}^{+2.2}$ & $\mathbf{43.4}^{+3.2}$ & $\mathbf{26.4}^{+2.3}$ & $\mathbf{6.2}^{+1.9}$ & $\mathbf{12.0}^{+1.4}$ & $\mathbf{22.9}^{+1.7}$ & $\mathbf{37.3}^{+2.6}$\\   \midrule
      \multirow{2}{*}{DOTA-v2.0}    & YOLOx   & 39.5 & 61.3 & 41.4 & 2.1	&9.4	&27.2	&\textbf{46.8} \\ 
                                    & YOLOx*  & $\mathbf{40.1}^{+0.6}$ & $\mathbf{63.0}^{+1.7}$ & $\mathbf{42.4}^{+1.0}$ & $\mathbf{2.3}^{+0.2}$ & $\mathbf{10.7}^{+1.3}$ & $\mathbf{31.3}^{+4.1}$ & $46.4^{-0.4}$\\   \bottomrule
  \end{tabular}}
  \end{table}

  \begin{table*}[hbtp]
  \center
  \caption{AI-TOD V2}
  \label{tab:tabel8}
  \renewcommand{\arraystretch}{1.5}
    \resizebox{0.8\textwidth}{!}
  {\begin{tabular}{ccccccccccc}
  \toprule
  Methods & Publication & \multicolumn{1}{c|}{Backbone} &  
  $AP$ & \textbf{$AP_{50}$}& \multicolumn{1}{c|}{$AP_{75}$} & $AP_{vt}$ & \textbf{$AP_{t}$ }& \textbf{$AP_{s}$} & \textbf{$AP_{m}$} \\ \midrule
  Two-Stage Detectors: & ~ & ~ & ~ & ~ & ~ & ~ & ~ & ~ & ~ & ~ \\ 
  TridentNet~\cite{TridentNet2019} & ICCV'2019 & ResNet-50 & 10.1  & 24.5  & 6.7  & 0.1  & 6.3  & 19.8  & 31.9  \\ 
  Faster R-CNN~\cite{FasterRCNN2017} & ArXiv'2015 & ResNet-50 & 12.8  & 29.9  & 9.4  & 0.0  & 9.2  & 24.6  & 37.0  \\ 
  Cascade R-CNN~\cite{CascadeRCNN2018} & CVPR'2018 & ResNet-50 & 15.1 & 34.2  & 11.2  & 0.1  & 11.5  & 26.7  & 38.5  \\ 
  DetectoRS~\cite{DetectoRS2021} & CVPR'2021 & ResNet-50 & 16.1  & 35.5  & 12.5  & 0.1  & 12.6  & 28.3  & 40.0  \\ 
  DotD~\cite{DotD2021} & CVPRW'2021 & ResNet-50 & 20.4  & 51.4  & 12.3  & 8.5  & 21.1  & 24.6  & 30.4  \\ 
  RFLA~\cite{RFLA2022} & ECCV'2022 & ResNet-50 & 25.7  & 58.9  & 18.8  & 9.2  & 25.5  & 30.2  & 40.2  \\ 
  CFINet~\cite{CFINet2023} & ICCV'2023 & ResNet-50  & ~ & ~ & ~ & ~ & ~ & ~ & ~ \\ 
  RepPoints~\cite{Reppoints2019} & ICCV'2019 & ResNet-50 & 9.3  & 23.6  & 5.4  & 2.8  & 10.0  & 12.3  & 18.9  \\ \midrule
  One-Stage Detectors:  & ~ & ~ & ~ & ~ & ~ & ~ & ~ & ~ & ~ \\ 
  YOLOv3~\cite{Yolov32018} & CoRR'2018 & Darknet-53 & 4.1  & 14.6  & 0.9  & 1.1  & 4.8  & 7.7  & 8.0  \\ 
  RetinaNet~\cite{RetinaNet2020} & ICCV'2017 & ResNet-50 & 8.9  & 24.2  & 4.6  & 2.7  & 8.4  & 13.1  & 20.4  \\ 
  SSD-512~\cite{SSD2016} & ECCV'2016 & ResNet-50 & 10.7  & 32.5  & 4.0  & 2.0  & 8.7  & 16.8  & 28.0  \\ 
  FCOS~\cite{FCOS2020} & ICCV2019 & ResNet-50 & 12.0  & 30.2  & 7.3  & 2.2  & 11.1  & 16.6  & 26.9  \\ 
  YOLOx~\cite{YOLOX2021} & ArXiv'2021 & CSPDarknet-53  & 23.7  & 55.2  & 16.4  & 9.0  & 22.8  & 29.8  & 40.1  \\ \hline
  Ours & ~ & CSPDarknet-53  & 26.3  & 58.2  & 20.4  & 13.7  & 24.3  & 33.5  & 44.8 \\ \hline
  \end{tabular}}
  \end{table*}
  
  \begin{table*}
  \center
  \caption{AI-TOD V2 per class}        
  \label{tab:tabel9}         
  \renewcommand{\arraystretch}{1.5}
    \resizebox{0.8\textwidth}{!}
  {\begin{tabular}{ccc|cccccccc}
  \toprule
  \textbf{Method} & \textbf{Publication} & \textbf{Backbone} & \textbf{AI} & \textbf{BR} & \textbf{ST} & \textbf{SH} & \textbf{SP}& \textbf{VE} & \textbf{PE} & \textbf{WM} \\ \hline
  Two-Stage Detectors: & ~ & ~ & ~ & ~ & ~ & ~ & ~ & ~ & ~ & ~ \\ 
  TridentNet~\cite{TridentNet2019} & ICCV'2019 & ResNet-50 & 19.3  & 0.1  & 17.2  & 16.2  & 12.4  & 12.5  & 3.4  & 0.0  \\ 
  Faster R-CNN~\cite{FasterRCNN2017} & ArXiv'2015 & ResNet-50 & 19.7  & 4.8  & 19.0  & 19.9  & 3.7  & 14.4  & 4.8  & 0.0  \\ 
  Cascade R-CNN~\cite{CascadeRCNN2018} & CVPR'2018 & ResNet-50 & 26.2  & 9.6  & 24.0  & 24.3  & 13.2  & 17.5  & 5.8  & 0.1  \\ 
  DetectoRS~\cite{DetectoRS2021} & CVPR'2021 & ResNet-50 & 28.5  & 11.7  & 23.2  & 26.4  & 14.9  & 17.6  & 6.5  & 0.2  \\ 
  DotD~\cite{DotD2021} & CVPRW'2021 & ResNet-50 & 18.7  & 17.5  & 34.7  & 37.0  & 12.4  & 25.4  & 10.3  & 7.4  \\ 
  RFLA~\cite{RFLA2022} & ECCV'2022 & ResNet-50 & ~ & ~ & ~ & ~ & ~ & ~ & ~ & ~ \\ 
  CFINet~\cite{CFINet2023} & ICCV'2023 & ResNet-50 & ~ & ~ & ~ & ~ & ~ & ~ & ~ & ~ \\ 
  RepPoints~\cite{Reppoints2019} & ICCV'2019 & ResNet-50 & 0.0  & 0.1  & 22.5  & 28.8  & 0.2  & 18.3  & 4.1  & 0.0  \\ 
  One-Stage Detectors: & ~ & ~ & ~ & ~ & ~ & ~ & ~ & ~ & ~ & ~ \\ 
  YOLOv3~\cite{Yolov32018} & CoRR'2018 & Darknet-53 & 0.3  & 0.5  & 8.5  & 9.4  & 0.0  & 12.7  & 1.4  & 0.0  \\ 
  RetinaNet~\cite{RetinaNet2020} & ICCV'2017 & ResNet-50 & 1.3  & 11.8  & 14.3  & 23.6  & 5.8  & 11.4  & 2.3  & 0.5  \\ 
  SSD-512~\cite{SSD2016} & ECCV'2016 & ResNet-50 & 14.9  & 9.6  & 13.2  & 18.2  & 10.6  & 12.7  & 2.9  & 3.1  \\ 
  FCOS~\cite{FCOS2020} & ICCV2019 & ResNet-50 & 7.2  & 13.4  & 20.2  & 26.7  & 8.4  & 16.3  & 3.5  & 0.0  \\ 
  YOLOx~\cite{YOLOX2021} & ArXiv'2021 & CSPDarknet-53 & 36.2  & 13.6  & 38.1  & 36.1  & 15.0  & 28.8  & 15.2  & 6.4  \\ \midrule
  Ours & ~ & CSPDarknet-53 & 41.7  & 17.0  & 38.8 & 38.4 & 20.0 & 29.8 & 16.8 & 7.8 \\ \bottomrule
  \end{tabular}}
  \end{table*}

  \subsection{Ablation Studies}
  Experiments are performed by gradually increasing or changing the modules of the baseline to demonstrate the effectiveness of the method, so as to validate the effectiveness of the proposed method.

  \subsubsection{Effectiveness of individual component}
  The Values in the Table Are Recorded in Percentage.
  \begin{table*}
    \center
    \caption{Ablation Study of Individual Effectiveness of $SA^2B$ and SFL on AI-TOD Dataset. The Values in the Table Are Recorded in Percentage.}
    \label{Ablation_yolov5}
    \renewcommand{\arraystretch}{1.5}
    \resizebox{\textwidth}{!}
      { \begin{tabular}{ccc|ccc|cccc|cccccccc}
        \toprule
        \multicolumn{1}{l}{YOLOx} & $SA^2B$ & SFL & $AP$            & $AP_{50}$     & $AP_{75}$     & $AP_{vt}$     & $AP_t$        & $AP_s$          & $AP_m$  &AI	&BR	&ST	&SH	&SP	&VE	&PE	&WM  \\ \midrule
        \checkmark       &      &    & 22.2          & 52.4         & 15.2         & 7.7           & 21.2          & 29.7          & 35.5 &33.8 	&12.2 	&37.1 	&33.6 	&15.9 	&27.3 	&13.5 	&4.2       \\
        \checkmark       & \checkmark    &     & 24.0        & 55.4          & 17.7          & 9.1           & 23.6          & 30.5          & 37.1   &37.2 	&14.3 	&39.2 	&36.2 	&18.0 	&28.1 	&15.3 	&3.7       \\
        \checkmark       &      & \checkmark   & 23.7          & 55.3          & 17.1          & 8.5           & 22.8          & 29.9          & 37.6   &37.0 	&13.7 	&37.6 	&35.8 	&16.7 	&28.7 	&14.7 	&5.4       \\
        \checkmark       & \checkmark    & \checkmark   & \textbf{26.6} & \textbf{59.1} & \textbf{20.8} & \textbf{10.0} & \textbf{26.0} & \textbf{33.7} & \textbf{39.9} &\textbf{38.9} &\textbf{16.5} 	&\textbf{40.3} 	&\textbf{40.8} 	&\textbf{22.1} 	&\textbf{29.7} 	&\textbf{16.9} 	&\textbf{7.9} 
        \\ \bottomrule
        \end{tabular}
      }
  \end{table*}
  \begin{table*}
    \center
    \caption{Ablation Results (i.e., Percentage of AP Scores) of Individual Effectiveness of $SA^2B$ and SFL on AI-TOD Dataset. }
    \label{Ablation_yolox}
    \renewcommand{\arraystretch}{1.5}
    \resizebox{\textwidth}{!}
      { \begin{tabular}{ccc|ccc|cccc|cccccccc}
        \toprule
        \multicolumn{1}{l}{YOLOv5} & $SA^2B$ & SFL & $AP$            & $AP_{50}$     & $AP_{75}$     & $AP_{vt}$     & $AP_t$        & $AP_s$          & $AP_m$  &AI	&BR	&ST	&SH	&SP	&VE	&PE	&WM  \\ \midrule
        \checkmark       &      &    &11.9 	&28.5 	&7.6 	&3.1 	&12.0 	&14.2 	&20.3 	&9.3 	&6.4 	&20.3 	&32.9 	&0.9 	&19.3 	&5.3 	&0.7\\
        \checkmark       & \checkmark    &     &27.4 	&61.5 	&20.5 	&9.4 	&26.5 	&35.6 	&44.7 	&42.6 	&17.8 	&41.4 	&38.0 	&23.8 	&31.7 	&14.7 	&9.6\\
        \checkmark       &      & \checkmark   &28.8 	&62.2 	&22.9 	&9.5 	&28.0 	&37.4 	&48.2 	&44.6 	&19.9 	&42.5 	&40.4 	&26.3 	&32.4 	&16.1 	&8.4\\
        \checkmark       & \checkmark    & \checkmark   &\textbf{29.6} 	&\textbf{62.6} 	&\textbf{24.1} 	&\textbf{11.4} 	&\textbf{28.7} 	&\textbf{38.6} 	&\textbf{47.4} 	&\textbf{44.5} 	&\textbf{19.7} 	&\textbf{43.3} 	&\textbf{42.0} 	&\textbf{26.9} 	&\textbf{33.7} 	&\textbf{16.2} 	&\textbf{10.6}\\ \bottomrule
        \end{tabular}
      }
    \end{table*}
  \subsubsection{Performance of different adjust factor}
  \begin{table*}[!htbp]
    \center
    \caption{Ablation Results (i.e., Percentage of AP Scores) of the Different Factor $\beta$ in the SFL on AI-TOD Dataset.}
    \label{Ablation_tab2}
    \renewcommand{\arraystretch}{1.5}
    \resizebox{0.9\textwidth}{!}
      { \begin{tabular}{c|ccc|cccc|cccccccc}
        \toprule
          $\beta$  & $AP$            & $AP_{50}$          & $AP_{75}$   & $AP_{vt}$          & $AP_t$           & $AP_s$          & $AP_m$   &AI	&BR	&ST	&SH	&SP	&VE	&PE	&WM \\ \midrule
          1              & 24.6          & 56.6               & 17.7        & 9.0                & 23.5             & 31.3            & 38.2     &37.9 &14.2 	&39.3 	&36.6 	&18.7 	&28.4 	&15.9 	&5.5         \\
          ${1}/{\ln 2}$  & 25.0          & 56.2               & 18.8        & 9.6                & 23.6             & 38.0            & 38.7     &37.6  &15.4 	&39.4 	&38.4 	&18.7 	&28.9 	&16.2 	&5.0         \\
          ${2}/{\ln 2}$  & \textbf{26.6} & \textbf{59.1} & \textbf{20.8} & \textbf{10.0} & \textbf{26.0} & \textbf{33.7} & \textbf{39.9}  &\textbf{38.9} &\textbf{16.5} 	&\textbf{40.3} 	&\textbf{40.8} 	&\textbf{22.1} 	&\textbf{29.7} 	&\textbf{16.9} 	&\textbf{7.9}        \\ \bottomrule
        \end{tabular}
      }
    \end{table*}

\section{Conclusion}\label{Conclusion}
In this paper,

\bibliographystyle{IEEEtran}
\bibliography{ref}

\end{document}